\documentclass[10pt,twocolumn,letterpaper]{article}

\usepackage[final]{cvpr}      
\usepackage[accsupp]{axessibility} 

\usepackage{graphicx}
\usepackage{amsmath}
\usepackage{amssymb}
\usepackage{booktabs}
\usepackage{multirow}

\usepackage[pagebackref,breaklinks,colorlinks]{hyperref}

\usepackage[capitalize]{cleveref}
\crefname{section}{Sec.}{Secs.}
\Crefname{section}{Section}{Sections}
\Crefname{table}{Table}{Tables}
\crefname{table}{Tab.}{Tabs.}

\def\cvprPaperID{28} 
\def\confName{CVPR}
\def\confYear{2022}

\title{V-Doc : Visual questions answers with Documents}

\author{
Yihao Ding\textsuperscript{1\thanks
{
co-first authors
}}, 
Zhe Huang\textsuperscript{1*},
Runlin Wang\textsuperscript{1}, 
YanHang Zhang\textsuperscript{1}, 
Xianru Chen\textsuperscript{1},\\ 
Yuzhong Ma\textsuperscript{1}, 
Hyunsuk Chung\textsuperscript{2}, 
Soyeon Caren Han\textsuperscript{1}
\thanks
{
Corresponding author (caren.han@sydney.edu.au)
}\\
\textsuperscript{1}The University of Sydney \textsuperscript{2}FortifyEdge \\
\tt\small 
\{yihao.ding, hyunsuk.chung, caren.han\}@sydney.edu.au,\\ 
\tt\small 
\{zhua8534, rwan2687, yzha5308, xche9640, yuma6895\}@uni.sydney.edu.au
}

\begin{document}
\maketitle 
\begin{abstract}
We propose V-Doc, a question-answering tool using document images and PDF, mainly for researchers and general non-deep learning experts looking to generate, process, and understand the document visual question answering tasks. The V-Doc supports generating and using both extractive and abstractive question-answer pairs using documents images. The extractive QA selects a subset of tokens or phrases from the document contents to predict the answers, while the abstractive QA recognises the language in the content and generates the answer based on the trained model. Both aspects are crucial to understanding the documents, especially in an image format. We include a detailed scenario of question generation for the abstractive QA task. V-Doc supports a wide range of datasets and models, and is highly extensible through a declarative, framework-agnostic platform.\footnote{Data and demo video: \url{https://github.com/usydnlp/vdoc}}
\end{abstract}

\section{Introduction}
\label{sec:intro}

Visual Question Answering (VQA) is a multi-modal deep learning to answer text-based questions about an image. There are a set of VQA tasks defined based on various application scenarios, including statistical charts \cite{methani2020plotqa,kahou2017figureqa}, daily-life photos \cite{hudson2019gqa} and digital-born documents \cite{mathew2021docvqa}. Among those VQA tasks, Document-VQA, which aims to extract information from documents and answer natural language questions, is more challenging. It requires detecting the scene objects and understanding the meaning and relation of those objects in order to predict reliable answers. Generally, existing QA can be classified into abstractive and extractive tasks. Extractive QA is to predict the answers by selecting a subset of tokens or phrases from document contents, while the answer of abstractive QA is generated by recognizing the document content based on the trained model. Different datasets and state-of-the-art models are introduced to deal with abstractive and extractive VQA problems.

\begin{figure}[t]
 \centering
 \includegraphics[width=\linewidth]{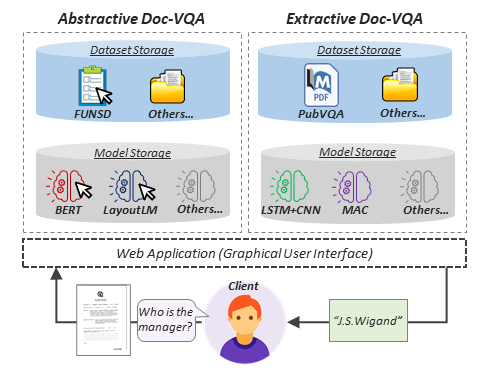}
 \caption{Overview of V-Doc platform architecture. Users can select sample images from datasets and use the trained models to predict the answer to input questions. Comparing the predicted answer between different models may represent the advantages of user proposed models to target clients.}
 \label{fig:system}
\end{figure}

\label{sec:system_architecture}

Recently, there are several Document-VQA datasets proposed by using website \cite{tanaka2021visualmrc}, textbook slides \cite{kembhavi2017you}, scanned documents \cite{jaume2019funsd} or cross-domain documents \cite{mathew2021docvqa}. Many recent VQA models widely use those datasets, but most datasets use only single page domain documents such as \cite{tanaka2021visualmrc,methani2020plotqa} which ignore the situation of more frequently used multi-page documents in real world. In addition, most of existing abstractive QA task only focus on charts or figures  \cite{methani2020plotqa,kahou2017figureqa} not whole document pages.

Transformer and BERT architectures \cite{devlin2019bert,liu2019roberta} have gained many remarkable achievements in many natural language tasks, including text classification, language generation and text question answering. VQA challenges was solved by different multi-modal models, such as bilinear models \cite{fukui2016bilinear}, image-text fusing based transformers \cite{xu2021layoutlmv2,li2019visualbert}, self-attention \cite{gao2018selfattention}, etc. Although there are many state-of-the-art models proposed recently, there is no platform that can represent the novel model performance and compare it with some classical baselines.

This paper proposes a platform, V-Doc, which can allow users to upload their trained abstractive or extractive models on different datasets to compare with other baselines to represent the model improvements and advantages to their clients or targeting users. This platform mainly contains three modules from designing Dataset Storage and Model Storage modules to demonstrate the performance by a web application. Dataset Storage module contains a Compositional Question Generation Engine to generate a new document VQA dataset based on multi-page documents collected from PubMed Open Source Subset\footnote{\href{https://pubmed.ncbi.nlm.nih.gov}{https://pubmed.ncbi.nlm.nih.gov}}. It also provides some well pre-processed, commonly used Document-VQA datasets such as FUNSD. User can add and train/test with more dataset to the systems. The Model Storage module provides some trained models on different datasets in our Dataset Storage module, which also provides an environment to support user uploading and testing user trained models in our platform. Our web application can directly use the trained models in the Model Storage module to show the availability and reliability of selected models to target clients.

The contributions of our paper mainly contain:
\begin{itemize}
    \item To the best of our knowledge, Our V-Doc is the first platform for Document-VQA which contains three modules including Dataset Storage, Model Storage and Graph User Interface for users. 
    \item We introduce a new dataset generation engine and introduce a new Document-VQA dataset, PubVQA which can be accessed in our Dataset Storage Module for filling the gap of lacking abstractive Document-VQA dataset in this area.
    \item An extendable Data and Model Storage module is provided to support user uploading and comparing their novel models with baselines through a user friendly graph user interface.
\end{itemize}

\section{System Architecture}
The V-Doc mainly contains three modules: 1) Dataset Storage, 2) Model Storage, and 3) a web application. The Dataset Storage provides existing pre-processed benchmark datasets and a question generation engine to generate QA pairs based on PubMed accessed papers. The Model Storage provides an environment to upload and store various trained Document-VQA models. The proposed web application can call the trained models based on users' demands.
 
The Dataset Storage module provides trainable pre-processed datasets and a new question generation engine to generate new QA pairs. The raw PDF documents are collected from PubMed. Each document is split into several images and fed into the pre-trained Mask RCNN\cite{zhong2019publaynet} model and Google vision API\footnote{\href{https://cloud.google.com/vision}{https://cloud.google.com/vision}} to generate the scene graphs of collected documents. The compositional question generation engine processes the scene graph files in order to generate a new abstractive Document-VQA dataset, named PubVQA. 

The classical and widely-used deep learning models, MAC\cite{hudson2018compositional}, LayoutLMv2\cite{xu2021layoutlmv2}, are included in the Model Storage module to deal with reading comprehension (abstractive) or information retrieval(extractive) tasks, respectively. Those typical architectures can be used as baselines to compare user uploaded trained models. This system allows user-defined or trained models built on commonly used machine learning frameworks.

The web application provides a graphical user interface for users instead of calling the predict method. The V-Doc web application consists of front-end designing and back-end development. The front-end is built by the JavaScript React framework, which is responsible for rendering the GUI, handling the input from users and passing it to the back-end, finally, rendering the answer that was collected from the back-end. Our back-end is built by the Python Flask restful framework. It loads the models from the Model Storage and calls the pre-defined predict functions via the machine learning framework, including Tensorflow and Pytorch. Each model is deployed in a separate back-end server to avoid the dependencies conflict. Thus, users can get the prediction from any particular model by uploading an image and writing down the question in the user interface.

\section{Dataset Storage}
\label{sec:dataset}
The Dataset Storage module provides some pre-processed Document-VQA datasets. It also contains a question generation module that provides a Compositional Question Generation Engine with well-processed scene graphs of collected PubMed papers to generate a new Document-VQA dataset, PubVQA. The question generation procedure mainly includes data collection, scene graph generation, and question generation.

\subsection{PubVQA Dataset Generation}
\label{sec:pubvqa}

\subsubsection{Data Collection}
\label{data_collection}
We collect a set of PDF documents from PubMed Open Access Subset based on PMCID and split the PDF documents into separate document images(.jpg file)\footnote{\href{https://github.com/Belval/pdf2image}{https://github.com/Belval/pdf2image}}. A pre-trained Mask RCNN model provided by IBM based on their proposed large-scale PubLaynet\cite{zhong2019publaynet} Dataset is used to conduct object detection for acquiring bounding box coordinates and the category type of each detected segment. Since the domain of PubLaynet is the same as our dataset, the outputs from this pre-trained Mask-RCNN model are reasonable and acceptable.

\subsubsection{Scene Graph Generation}
After acquiring results from Mask-RCNN, we need to generate the scene graph of each image for representing the relation between detected segments. The generated scene graph can provide essential information and be used by the question generation engine to generate QA pairs of PubVQA datasets. The key attributes in the scene graph files contains \textit{text content}, \textit{reading order}, \textit{parent child relation}, \textit{relative position}, etc. The procedures of acquiring those attributes to generate scene graph files include:
 
\begin{figure}[ht]
    \hspace*{0cm}
     \centering
     \begin{subfigure}[b]{0.2\textwidth}
         \centering
         \includegraphics[width=3.71cm]{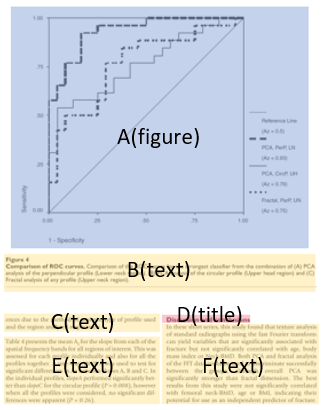}
         \caption{original image}
         \label{fig:original}
     \end{subfigure}
     \hspace*{0em}
     \begin{subfigure}[b]{0.2\textwidth}
         \centering
         \includegraphics[width=3.8cm]{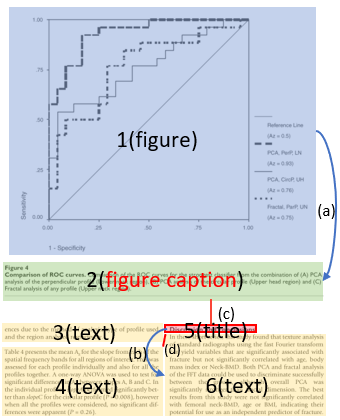}
         \caption{Pre-processed page}
         \label{fig:preprocessed}
     \end{subfigure}
        \caption{Scene Graph Generation Preprocessing Examples}
        \label{fig:scene graph}
\end{figure}

\begin{itemize}

\item\textbf{Content Extraction:} 
We apply the Google Cloud Vision API to implement the Optical Character Recognition (OCR) to extract the text content of each PDF segment (bounding box). 

\item\textbf{Reading Order:} 
We design a method for assigning an index number to each segment according to the reading order that people tend to do. It mainly contains two steps: (1) Ordering segments based on the top-left y-coordinate to generate an initial Reading Order List. (2) If any segment has neighbours on its left side, but the current Reading Order List index is larger than that segment, exchange the location in Reading Order List. Keep doing step 2 to generate the final Reading Order List. Base on the proposed methods, the reading order of segment $A$ to $F$ in Figure \ref{fig:original} is \textit{A(figure), B(text), C(text), E(text), D(title), F(text)} and the index of each segment is represented in Figure \ref{fig:preprocessed}.

\item\textbf{Gap Distance:} 
We obtain the gap distance by calculating the shortest distance between two bounding boxes. There are three rules: 
(1) If two bounding boxes overlap, the gap distance between the boxes is counted as 0. 
(2) If both bounding boxes contain the sides that partially intersect on the x-axis or y-axis direction, the gap distance is the vertical or horizontal distance between two sides. 
(3) If two bounding boxes have no intersection on the x-axis and y-axis, the gap distance is between the two closest vertices. The line segment \textbf{(c)}, \textbf{(d)} in Figure \ref{fig:preprocessed} represent the gap distance between segment \textit{D(title)} to segment \textit{B(text)} and \textit{E(text)}, respectively.

\item\textbf{Category Refinement:} 
We have five categories including $text$, $title$, $list$, $table$, $figure$ for the outputs from pre-trained Mask-RCNN on Publaynet. However, in order to generate more comprehensive QA pairs, the predicted category of some segments should be updated especially for altering some $text$ segments to $table$ $caption$ or \textit{figure caption}. Most of table and figure captions are the segments closest to the table or figure. In this case, based on the calculated gap distance results, it is easy to change the closest $text$ segment to a $table$ or $figure$ segment to \textit{table caption} or \textit{figure  caption}. Based on those rules, the category of segment $B$ in Figure \ref{fig:original} is updated from $text$ to \textit{figure caption} in Figure \ref{fig:preprocessed}.

\item\textbf{Parent Child Relation:} 
We easily generate the parent-child relation between document segments by using the reading order and category refinement results. There are three rules: 
(1) If there is not other $title$ objects between two $title$ objects based on the Reading Order List, the first title is the parent of all $text$ or $list$ segments until the next $title$ appearing such as relation \textbf{(b)} in Figure \ref{fig:preprocessed} . 
(2) If there is no other types of segments between them based on reading order sequence, $text$ or $title$ segment is the parent of the $list$ segments. 
(3) If there is a \textit{table caption} or \textit{figure caption}, $table$ or $figure$ is the parent of the nearest caption segment. For example, based on Figure \ref{fig:preprocessed} relation \textbf{(a)}, segment \textit{A(figure)} is the parent of segment \textit{B(figure caption)}. 

\item\textbf{Relative Position:}
We assign the relative position between two segments is assigned based on their bounding box coordinates. For example, in Figure \ref{fig:preprocessed}, if two segments coincide in the y-axis direction, but not in the x-axis like relative position between element $C$ and $D$. We can define $C$ as on the left of $D$, and $D$ is on the right of $C$. If two segments do not coincide in both x and y-axis direction (e.g. between segments $E$ and $D$), we can define that $D$ is located on the right-top of $E$ and $E$ is on the left-bottom of $D$. 
\end{itemize}

 \begin{table*}[t]
   \caption{Example of questions and answers for each template}
   \label{table:template}
   \centering
   \begin{tabular}{ lll }
     \toprule
     \textbf{Question Type} & \textbf{Sample Question Template} & \textbf{Answers Type} \\
     \midrule
     \multirow{2}{*}{Position Related} & Where is the caption of the $<$E$>$ located at? & top/bottom\\
     & The caption is on which side of the $<$E$>$? & top/bottom \\
     \midrule
     \multirow{2}{*}{Counting Based} & How many $<$E1$>$ objects are located at the $<$R$>$ side of $<$E2$>$? & number \\
     & For $<$E2$>$, how many $<$E1$>$ objects are located at its $<$R$>$ side? & number \\
     \midrule
     \multirow{2}{*}{Existence Based} & Do title objects exist on this page? & yes/no \\
     & Are there any titles that exist? & yes/no\\
     \bottomrule
 \end{tabular}
 \end{table*}
 
\subsubsection{Question generation}
Inspired by the Clevr\cite{johnson2017clevr} proposed question generation producer on synthetic images, we designed a Compositional Question Generation Engine for real-world PDF documents by using generated scene graph files of each document image. Similar to Clevr question generation methods, some templates and functional programs are defined to generate QA pairs. Table \ref{table:template} shows the three different groups of templates, including position related, counting based and existence based with expected answer type. 

\begin{figure}[ht]
 \centering
 \includegraphics[width=\linewidth]{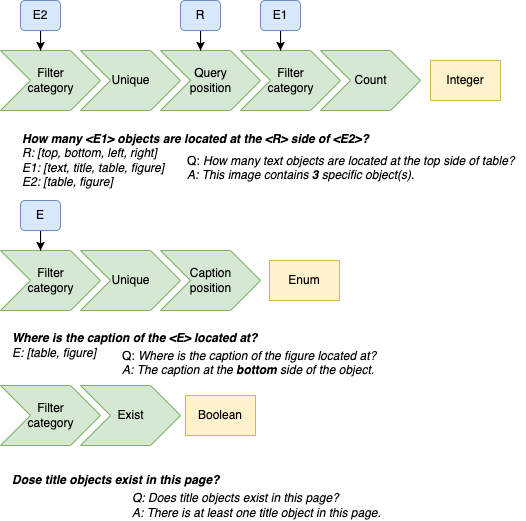}
 \caption{Position related, counting based and existence based question generation pre-defined function sequences}
 \label{fig:qageneration}
\end{figure}

Figure \ref{fig:qageneration} represents the workflows of the sequences of pre-defined functions to generate the corresponding QA pairs. For example, the position-related question workflow mainly consists of three pre-defined functions. To generate a question, \textit{'Where is the caption of the $<$E$>$ located at ?'}, the engine needs to filter the target segments by using $Filter$ $category$ function and then pass through a $Unique$ function to select one target $table$ or $figure$ segment randomly. Then, the \textit{Caption position} function can extract the relative location between target $table$ and its caption.

\begin{table}[h]
\caption{Existing Dataset size (question-answer pairs)}
\centering
\begin{tabular}{ llll } 
   \midrule
   Dataset & Training & Testing & Validation \\ 
   \midrule
   PubVQA & 1260 & 360 & 180 \\ 
   FUNSD-QA  & 1000 & 300 &  \\ 
   \bottomrule
 \end{tabular}
 \label{tab:dataset}
 \end{table}
 
\subsection{Existing Dataset}
The V-Doc have two following default datasets, which are publicly available and widely used datasets in the Document VQA domain.
\begin{itemize}
\item\textbf{FUNSD-QA:} 
It is a form understanding dataset selected from RVL-CDIP \cite{harley2015evaluation} dataset. It is created for scanned document-related tasks, including text detection, OCR, layout analysis. It contains 199 PDF images and provides the category of layout component, bounding box coordinates, text content and links between entities. Based on the pre-defined four category types ('Question', 'Answer', 'Header', 'Other') and the links between 'Question' and 'Answer', we can easily generate a span classification based extractive QA dataset named FUNSD-QA. 

\item\textbf{PubVQA:} 
We introduce/create a new abstractive Document-VQA dataset, named PubVQA. The dataset split information is list in Table \ref{tab:dataset}. Some classical abstractive QA models are used on this dataset to generate trained models uploaded to our Model Storage module.
\end{itemize}

\section{Model Storage}
The V-Doc provides a Model Storage Module to manage existed or user-uploaded trained models. It provides the environment to support those models built on commonly used machine learning frameworks. Similar to Dataset Storage module, the existing trained model in current Model Storage module also can be classified into extractive and abstractive types. Many pre-trained language models can be fine-tuned on spanning based question answering tasks to extract answer from the input texts by predicting the start and end index such as BERT \cite{devlin2019bert}, LayoutLMv2 \cite{xu2021layoutlmv2}. Those models are trained on extractive datasets provided by the Dataset Storage module which can be used as baselines and compared with novel proposed models. In addition, some widely used abstractive baselines are also provided in Model Storage module which can be directly used to generate the predicted answers and compared with user uploaded models.

\subsection{Extractive QA Models}
The V-Doc has two extractive QA Models, included in the model storage.
\begin{itemize}
\item\textbf{BERT}: 
It is a widely used multi-layer bi-directional transformer-based language model. It can be easily applied to solve span classification based question answering models and trained on QA datasets, such as SQuAD (The Standford Question Answering Dataset). The linear layers are placed on the top of pre-trained BERT to predict the start and end logits for extracting the content from inputs. We fine-tune BERT-based QA models on the FUNSD-QA dataset to acquire a trained model uploaded to the Model Storage module.

\item\textbf{LayoutLMv2}: 
It is a multi-modal transformer-based architecture that not only is pre-trained by masked language modelling task but also some new text-image alignment and matching tasks are applied to learn the cross-modality interaction. It also provides a LayoutLMv2FeatureExtractor to extract the image features for pre-training and downstream tasks. Similar to BERT for extractive QA, a linear layer is placed on the top of hidden state outputs to predict the start and end index. A trained LayoutLMv2 model on the FUNSD dataset is provided in our Model Storage module too.
\end{itemize}

\subsection{Abstractive QA Models}
The V-Doc has two abstractive QA Models, included in the model storage.
\begin{itemize}
\item\textbf{LSTM+CNN}: It is a baseline model architecture adopted by MAC \cite{hudson2018compositional}. It uses Long Short Term Memory (LSTM) network to encode textual information of question, and uses pre-trained convolution neural network to learn visual features of document image. Then, the concatenated question and visual embeddings are fed into a linear layer to get the predicted answer logits. 

\item\textbf{MAC}: It is an end-to-end differentiable architecture which contains sequence of MAC cells to perform the multi-stage reasoning process. Each MAC recurrent cell contains three units including control unit, read unit and write unit to manage, extract and integrate question and image features separately. MAC network has achieved state-of-the-art performance on Clevr \cite{johnson2017clevr} dataset, which is the reason why we choose this model as one our provided baseline model in Model Storage module. 
\end{itemize}

\section{Web Application}

\subsection{Graphical User Interface}
The user interface of the V-Doc system is built as a single page Web Application by using Nodejs and React framework. All components and icons are imported from Semi design\footnote{\hyperlink{https://semi.design/en-US/start/introduction}{https://semi.design/en-US/start/introduction}}. Mainly, this Web Application sends the PDF images and questions to the server with different prediction models and shows the result in the table below. The GUI of the V-Doc system mainly consists of four sections: model selection, image selection, input question and answer representation.

\begin{figure}[ht]
 \centering
 \includegraphics[width=\linewidth]{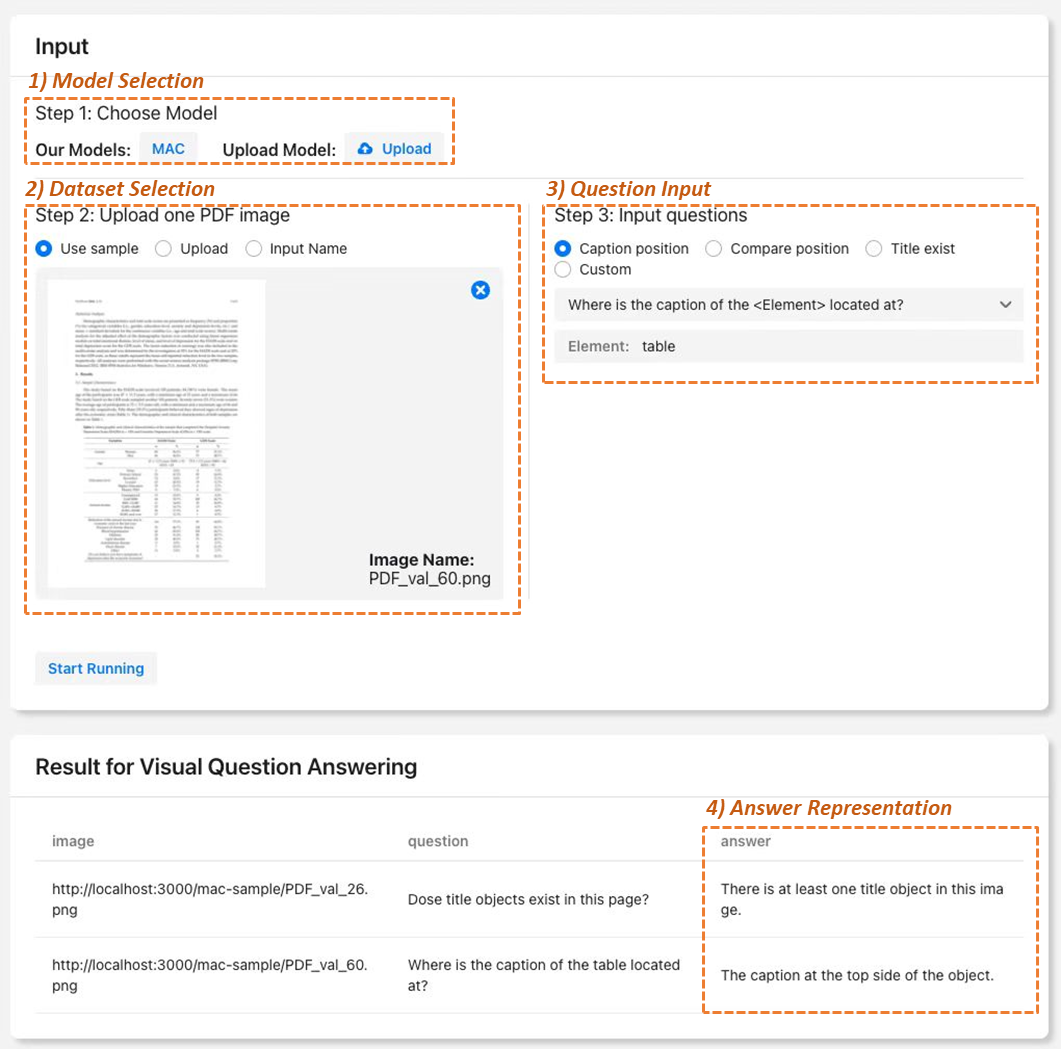}
 \caption{Interface of Web Application. The key subsections are highlighted including Model Selection, Question Input, Dataset Selection and Predicted Answer Representation}
 \label{fig:input}
\end{figure}

\textbf{Model selection: } It allows users to select trained models provided by Model Storage module.
\textbf{Dataset selection: } It provides a image selector that normally contains nine PDF images, including six sample images from PubVQA dataset and three images from FUNSD-QA dataset. Users can select a sample image, and the system can directly apply the chosen model and file it into the input stage. Each sample indicates the template and a recommended input question. Users are allowed to use other images stored in Dataset Storage module in this Dataset selection section.

\textbf{Question Input: }As shown in Figure \ref{fig:input}, our V-Doc system accepts both pre-defined templates and custom questions. When a pre-defined template is chosen, question will be composed by skeleton and parameters. Each model has several particular templates, as the suggested questions. After the question is decided, selected trained model, sample image and question will be sent into back-end to generate predicted answer.

\textbf{Answer Representation: }When the result returned from back-end, the front-end will rendering the file name, question and predicted answer into result table. The results will be saved until user leaves the page.

\begin{figure}[ht]
 \centering
 \includegraphics[width=\linewidth]{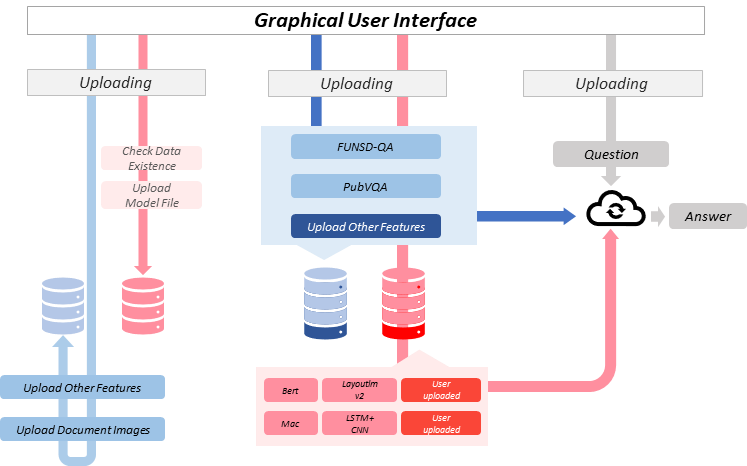}
 \caption{User oriented workflow from model and dataset uploading, trained model and dataset selection, question inputting to get predicted answer.}
 \label{fig:userworkflow}
\end{figure}

\subsection{Backend Dependencies}
The dependencies of the V-Doc web application inherit from both provided MAC and LayoutLMv2. Both $torch 1.8.0$ and $tensorflow 1.15$ are implemented to adapt various models. The packages $imageio 2.9.0$ and $PIL 8.4.0$ are called to process the uploaded image, as a binary data. Then, $dataset 1.12.1$ and $pandas 1.3.4$ are used to collect and process both image data and uploaded question, preparing for the model pre-processing step. We applied $transformer 4.12.3$ build-in methods to pre-process and predict answer by trained models. Finally, $flask 2.0.2$, $flask-cors 3.0.10$ are used for building the backend to receive and response the request from front-end.

\subsection{User Instruction}
The interaction between users and the V-Doc GUI mainly contains three key steps. The first step is an optional one is to upload user trained model and dataset to Dataset Storage module or Model Storage module base on user demands. For uploading new dataset to Dataset Storage module, user should upload document images and other required embedding sequentially. Regarding to upload trained model to Model Storage module, user need to check whether the dataset existed in Dataset Storage before uploading their trained model. V-Doc GUI also allow users to select the existing trained models on provided datasets in Model Storage and Dataset Storage modules and send them to back-end of Web application. Then, users can input question and get the predicted answers based on selected dataset and model.

\section{Evaluation}
\subsection{Extractive QA performance}

\begin{table}[h]
\caption{Extractive QA model performance on FUNSD-QA}
\centering
\begin{tabular}{ lll } 
   \midrule
   \textbf{Model} & \textbf{BLUE}\\ 
   \midrule
   BERT  & 9.37  \\ 
   LayoutLMv2 & 11 \\ 
   \bottomrule
 \end{tabular}
 \label{tab:extractive_result}
 \end{table}

For solving extractive problems, we provide two trained model on FUNSD-QA dataset to predict the start index and end index of input documents. After getting the start and end index, we extract the actual tokens and calculate the average sentence-level BLUE score between ground-truth and predicted answers. The first provided Model is fine-tuned on pre-trained "bert-base-uncased" ignoring the visual aspect features of document images to provide a text-feature only baseline for comparing with other models. In addition, we also provided another multi-modal baseline which is fine-tuned on "microsoft/layoutlmv2-base-uncased". We use LayoutLMv2 provided Processor to encode multi-aspect features including text, bounding box, visual features. The BLUE score of two trained models are 9.37 and 11, respectively, which represents multi-aspect features can effectively improve the extractive QA model performance.
 
 \begin{table}[h]
\caption{Abstractive QA model performance on PubVQA dataset}
\centering
\begin{tabular}{ lll } 
   \midrule
   \textbf{Model} & \textbf{Val-Acc} & \textbf {Test-Acc}\\ 
   \midrule
   LSTM+CNN & 45.69 & 48.78  \\ 
   MAC &54.32  &55.44 \\
   \bottomrule
 \end{tabular}
 \label{tab:extractive_result}
 \end{table}
 
\subsection{Abstractive QA performance}
For anstractive task, we selected two commonly used baselines and tested on our PubVQA dataset including LSTM+CNN, MAC. We use one bi-directional LSTM layer (dim = 128) and extract the last hidden state, which is concatenated with the image visual feature (dim=1024) extracted from ResNet101 (same as MAC model adopted). For the MAC network, the only change is the output layer dimension is change to 9 which is equal to the length of PubVQA answer set.From Table \ref{tab:extractive_result}, MAC model can have better performance 55.44, than LSTM+CNN structure 48.78 on test dataset.





\section{Conclusion}
This paper introduces a new document visual question answering platform, V-Doc, that provides a web application to predict corresponding answers of input questions by using an uploaded trained model in the model storage. Currently, the extendable model storage contains both abstractive and extractive trained document related VQA baseline models on different datasets. The model storage can also allow a user to upload other trained models to demonstrate the performance in the web application. In addition, we also introduced a new PubVQA dataset generated by real-world medical journals by an extendable question generation engine. In summary, our system provides a platform for deep learning researcher to represent their VQA model results to fill the gaps in this area. In the future, we will update our model storage to provide more trained baseline models, and more question generation templates will be provided to improve the quality of our dataset.

{\small
\bibliographystyle{ieee_fullname}
\bibliography{egbib}
}

\end{document}